\documentclass[11pt]{article}

\usepackage[utf8]{inputenc}
\usepackage[T1]{fontenc}
\usepackage{amsmath,amssymb}
\usepackage{graphicx}
\usepackage{booktabs}
\usepackage{hyperref}
\usepackage{xcolor}
\usepackage{microtype}
\usepackage{natbib}
\usepackage{algorithm}
\usepackage{algorithmic}
\usepackage[margin=1in]{geometry}
\usepackage{caption}
\usepackage{subcaption}
\usepackage{enumitem}

\newcommand{\swerValue}{0.0054}
\newcommand{\swerRandom}{0.7778}
\newcommand{\swerRelative}{0.0069}
\newcommand{\rawAccuracy}{99.0\%}
\newcommand{\familyAccuracy}{99.4\%}

{--}
  \newcommand{\swerRandom}{--}
  \newcommand{\swerRelative}{--}
  \newcommand{\rawAccuracy}{--}
  \newcommand{\familyAccuracy}{--}
}

\hypersetup{
    colorlinks=true,
    linkcolor=blue,
    citecolor=blue,
    urlcolor=blue
}

\title{Parameter-Efficient Fine-Tuning of DINOv2 for\\Large-Scale Font Classification}

\author{
    Daniel Chen\thanks{Corresponding author: \texttt{daniel@anything.com}} \quad
    Marcus Lowe\thanks{\texttt{marcus@anything.com}} \quad
    Zaria Zinn\thanks{\texttt{zaria@anything.com}}
}

\date{}

\begin{document}
\maketitle

\begin{abstract}
We introduce GoogleFontsBench, the first public benchmark for classifying open-source web fonts, addressing a gap left by existing benchmarks that cover only commercial typefaces. GoogleFontsBench comprises 394 font variants across 32 Google Fonts families, a reproducible synthetic data generation pipeline (${\sim}$575 images per variant, ${\sim}$226K total), and a typographically-grounded evaluation metric (SWER) that weights errors by visual severity. We establish baselines using six fine-tuning strategies on a DINOv2 Vision Transformer backbone. Parameter-efficient adaptation with LoRA achieves 99.0\% top-1 accuracy while training only 1\% of the model's 87.2M parameters, with errors 140$\times$ less severe than random guessing. We release the benchmark, all trained models, and the full training pipeline as open-source resources.
\end{abstract}

\section{Introduction}

Font identification is a practical problem encountered across graphic design, document analysis, brand compliance, and web development. Given an image of rendered text, the task is to classify the typeface used. While humans with typographic expertise can distinguish common font families, the proliferation of fonts---Google Fonts alone hosts over 1,700 families with numerous stylistic variants---makes manual identification increasingly impractical.

Prior approaches to font recognition have relied on hand-crafted features such as stroke width, serif detection, and character-level shape descriptors \citep{zramdini1998optical}. More recent methods leverage convolutional neural networks trained on synthetic font images \citep{wang2015deepfont}. However, these systems often struggle with large label spaces, subtle inter-class differences (e.g., weight variants of the same family), and the domain gap between synthetic training data and real-world images.

Existing font recognition benchmarks such as AdobeVFR \citep{wang2015deepfont} focus almost exclusively on commercial typefaces. No public benchmark covers the open-source fonts that dominate modern web and mobile design---Google Fonts alone powers over 70 billion monthly font views. This gap leaves practitioners without a standardized way to evaluate font classification systems for the fonts they actually encounter.

In this work, we address both the benchmark gap and the classification task. We introduce \textbf{GoogleFontsBench}, a public benchmark for open-source font classification comprising 394 font variants across 32 families, a reproducible synthetic data generation pipeline, and a typographically-grounded evaluation protocol. We establish baselines on this benchmark using six fine-tuning strategies, finding that parameter-efficient adaptation of DINOv2 \citep{oquab2024dinov2} with LoRA \citep{hu2022lora} achieves the strongest efficiency--accuracy trade-off.

Our contributions are as follows:
\begin{enumerate}[leftmargin=*]
    \item \textbf{GoogleFontsBench}: the first public font classification benchmark for open-source web fonts, with 394 classes, ${\sim}$225K training images, and a standardized evaluation protocol including a typographic-distance-based severity metric (SWER).
    \item Baseline results from six fine-tuning strategies (linear probe, LoRA at three ranks, full fine-tuning, and ResNet-50), establishing reference points for future work.
    \item A parameter-efficient approach using LoRA on DINOv2 that achieves 99.0\% top-1 accuracy while training only 1\% of parameters, with errors 140$\times$ less severe than random guessing as measured by SWER.
    \item An end-to-end system with built-in preprocessing that ensures consistency between training and deployment, released as open-source on HuggingFace.
\end{enumerate}

This model is deployed as a component of a production design tool.\footnote{\url{https://anything.com}}

\section{Related Work}

\subsection{Font Recognition}

Early font recognition systems operated on scanned documents and relied on global texture features, typographic measurements, and template matching \citep{zramdini1998optical}. \citet{wang2015deepfont} introduced DeepFont, a CNN-based system trained on synthetic font images with domain adaptation to handle real-world photographs. Subsequent work explored multi-task learning for joint font attribute prediction \citep{chen2014large} and attention mechanisms for discriminating fine-grained typographic features.

\subsection{Self-Supervised Vision Transformers}

Vision Transformers (ViTs) \citep{dosovitskiy2021image} have become a dominant architecture in computer vision. Self-supervised pre-training methods---including DINO \citep{caron2021emerging} and its successor DINOv2 \citep{oquab2024dinov2}---learn powerful visual representations through self-distillation without labeled data. DINOv2 features emerge that capture semantic structure, texture, and fine-grained visual distinctions, making it well-suited as a backbone for downstream classification tasks.

\subsection{Parameter-Efficient Fine-Tuning}

Full fine-tuning of large pre-trained models is computationally expensive and risks catastrophic forgetting. Parameter-efficient methods such as adapter layers \citep{houlsby2019parameter}, prompt tuning \citep{jia2022visual}, and Low-Rank Adaptation (LoRA) \citep{hu2022lora} inject a small number of trainable parameters while keeping the backbone frozen. LoRA decomposes weight updates into low-rank matrices, enabling efficient adaptation with minimal overhead. It has been widely adopted for language models and has shown strong results on vision tasks as well.

\section{Dataset}
\label{sec:dataset}

\subsection{Font Selection}

We source fonts from the Google Fonts open-source repository.\footnote{\url{https://github.com/google/fonts}} From the full catalog, we select a curated set of 32 base font families spanning diverse typographic categories: sans-serif (e.g., Inter, Roboto, Poppins, DM Sans), serif (e.g., Crimson Pro, Playfair Display, Merriweather), monospace (JetBrains Mono), and display faces (e.g., Ultra, Big Shoulders Text). Selection was guided by three criteria: (1)~\emph{prevalence in modern design tooling}---we selected fonts commonly encountered in Figma, Dribbble, and similar platforms, based on the authors' experience building production design tools for website and mobile design; (2)~\emph{visual distinctiveness}---we excluded near-duplicate families that differ only in minor metrics to keep classes well-separated; and (3)~\emph{representativeness of high-quality digital design}---the set reflects the typographic choices typical of professional web and mobile interfaces. Many of these families include variable font axes (primarily weight), and we enumerate each axis variation as a separate class. This produces a total of 394 distinct font variants for classification.

\subsubsection{Iterative Font Curation}

The final font set is the result of four dataset iterations, each informed by the failures of the previous version. An initial proof-of-concept with 10~families (102~variants) achieved high accuracy but covered too few fonts to be practical. Expanding aggressively to over 700~variants from the broader Google Fonts catalog introduced many visually near-identical pairs (families differing only in minor spacing or hinting), which degraded accuracy substantially. Scaling back to 20~families helped but exposed insufficient training data and residual visual overlaps. The current version rebuilds the font set from scratch using the selection criteria described above, settling on 32~families with 394~variants and approximately 575~training images per variant. A critical engineering fix in this iteration---embedding preprocessing into the model's \texttt{forward} method---eliminated train--serve skew that had silently hurt earlier versions. All dataset versions are publicly available on HuggingFace.

\subsection{Synthetic Image Generation}

We generate training images by rendering text strings in each font variant onto images. The generation pipeline proceeds as follows:

\begin{enumerate}[leftmargin=*]
    \item \textbf{Text selection.} For each font variant, we render 500 random sentences drawn from literary source texts, plus 25 random numbers, 25 dollar amounts, and 25 percentage values. Test images use 25 sentences, 5 numbers, 5 dollar amounts, and 5 percentage values per class.

    \item \textbf{Rendering.} Text is rendered at 1024px font size onto a canvas with padding on all sides (128px) using the Pillow library. The canvas is then cropped to the tight bounding box of the rendered glyphs (i.e., all empty space beyond the ink is removed). The cropped glyph image is resized so that its height equals 224 pixels, with the width scaled proportionally to preserve the aspect ratio, using Lanczos resampling.

    \item \textbf{Color augmentation.} Background and text colors are randomly sampled as RGB tuples. To ensure legibility, we require that their perceived luminance values differ by at least 80 on a 0--255 scale, where luminance is computed as $L = 0.299R + 0.587G + 0.114B$ (the ITU-R BT.601 formula). If the randomly drawn pair fails this contrast check, a new pair is sampled until the constraint is satisfied.

    \item \textbf{Layout augmentation.} Text alignment is randomly chosen from left, center, or right. With 20\% probability, spaces are replaced by newlines to create multi-line renderings.

    \item \textbf{Noise augmentation.} Per-pixel Gaussian noise is added to improve robustness to compression artifacts and sensor noise. Each pixel channel value $p \in [0, 255]$ is replaced by $\mathrm{clamp}(p + \mathcal{N}(0,\, \sigma^2),\; 0,\; 255)$ where $\sigma = 0.10 \times 255 = 25.5$. In practice this means most pixel values are perturbed by roughly $\pm$25 intensity levels.
\end{enumerate}

The resulting dataset, \texttt{font\_crops\_v5},\footnote{\url{https://huggingface.co/datasets/dchen0/font_crops_v5}} contains approximately 575 training images per font variant, with a held-out test set of 40 images per variant.

\subsection{Data Validation}

All generated images are verified programmatically using PIL's image verification to detect and remove corrupted files. Filenames are sanitized to handle special characters that may cause filesystem issues across platforms.

\begin{figure}[t]
    \centering
    \includegraphics[width=\textwidth]{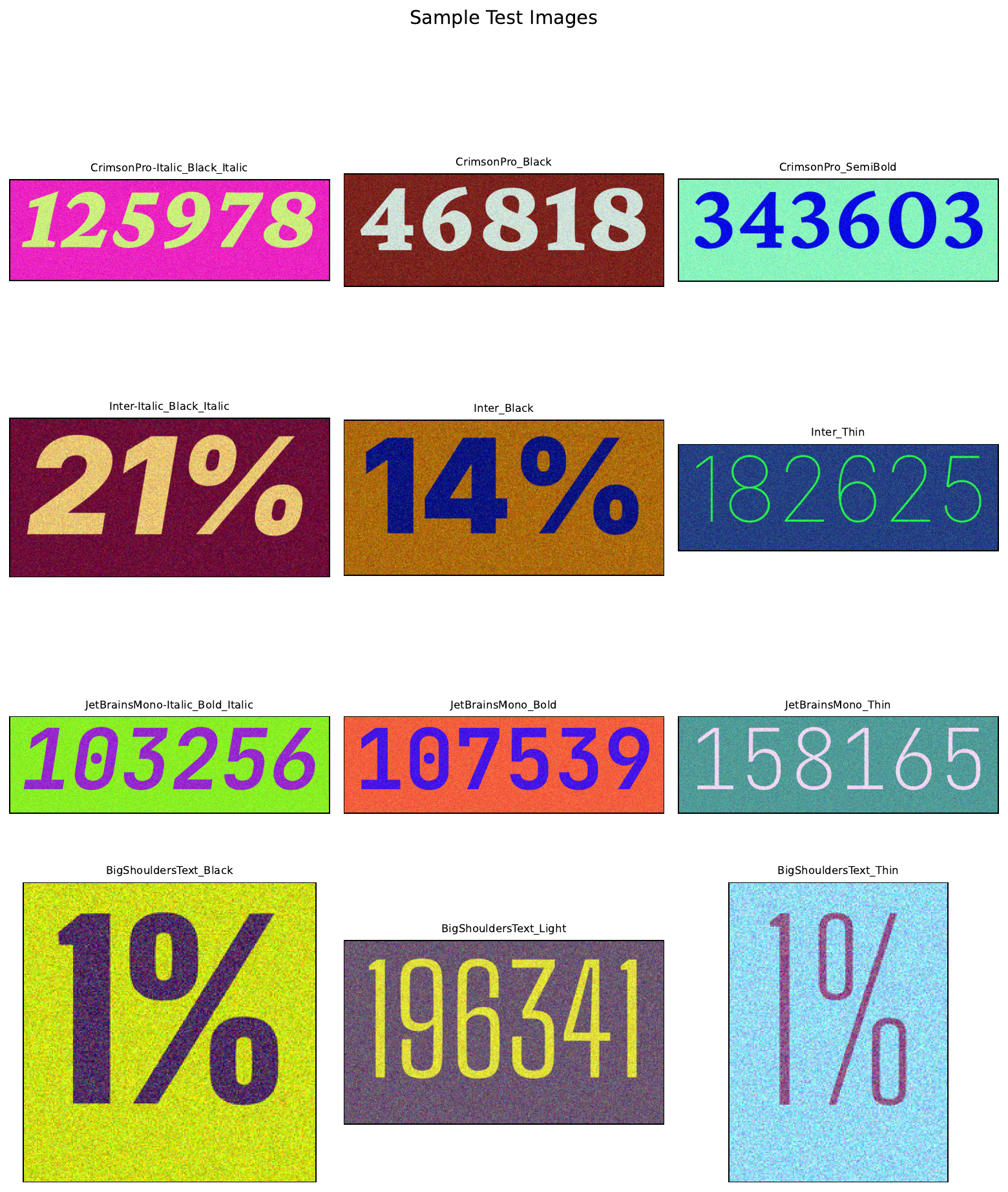}
    \caption{Sample test images from four font families spanning different typographic categories: serif (Crimson Pro), sans-serif (Inter), monospace (JetBrains Mono), and display (Big Shoulders Text). Each column shows a different weight variant. These are the actual rendered images seen by the model at inference time, including color augmentation and noise.}
    \label{fig:sample_images}
\end{figure}

\section{Method}

\subsection{Model Architecture}

Our model builds on the DINOv2-Base architecture \citep{oquab2024dinov2}, specifically the \texttt{dinov2-base-imagenet1k-1-layer} variant pre-trained on ImageNet-1K. This model uses the standard ViT-B/14 architecture with patch size 14, embedding dimension 768, 12 attention heads, and 12 transformer blocks.

We add a linear classification head mapping the \texttt{[CLS]} token representation to the 394 output classes. The classification head is randomly initialized and fully trainable.

\subsection{LoRA Configuration}
\label{sec:lora}

We apply LoRA to the query ($W_Q$) and value ($W_V$) projection matrices in each transformer block's multi-head self-attention layer. The LoRA decomposition replaces a weight update $\Delta W \in \mathbb{R}^{d \times d}$ with the product of two low-rank matrices:
\begin{equation}
    \Delta W = BA, \quad B \in \mathbb{R}^{d \times r}, \quad A \in \mathbb{R}^{r \times d}
\end{equation}
where $r \ll d$ is the rank. We use $r = 8$ and scaling factor $\alpha = 16$. The scaling factor controls the magnitude of the low-rank update: the actual weight modification applied is $\frac{\alpha}{r} \Delta W$, so with $\alpha = 16$ and $r = 8$ the update is scaled by a factor of 2. Dropout with probability 0.1 is applied to the low-rank matrices $A$ and $B$ during training as regularization. The base model weights remain frozen; only the LoRA parameters and the randomly initialized classification head are updated.

This configuration adds approximately 295K trainable LoRA parameters. Together with the 606K-parameter classification head (which is also trainable), the total number of trainable parameters is approximately 900K, or roughly 1\% of the 87M frozen backbone.

\subsection{Preprocessing}

A critical design decision is ensuring identical preprocessing at training and inference time. Our preprocessing pipeline:
\begin{enumerate}[leftmargin=*]
    \item Convert to RGB.
    \item Pad to a square canvas: since the rendered glyph images are typically wider than they are tall, black pixels (\texttt{fill=0}) are added equally to the top and bottom (or left and right) so that the shorter dimension matches the longer one. The image is centered within the padded square.
    \item Resize the square image to $224 \times 224$ pixels.
    \item Convert to tensor and normalize using ImageNet statistics (mean $= [0.485, 0.456, 0.406]$, std $= [0.229, 0.224, 0.225]$).
\end{enumerate}

To eliminate preprocessing discrepancies, we define a shared transform pipeline (\texttt{get\_inference\_transform} in \texttt{handler.py}) that is applied before every model call. The \texttt{EndpointHandler} class used for HuggingFace Inference Endpoints applies this same pipeline, ensuring that any deployment path---inference endpoint, batch evaluation, or direct model call---uses identical transformations.

\section{Experimental Setup}

\subsection{Training Details}

We train using the HuggingFace \texttt{Trainer} with the following hyperparameters:

\begin{table}[h]
\centering
\caption{Training hyperparameters.}
\label{tab:hyperparams}
\begin{tabular}{ll}
\toprule
\textbf{Hyperparameter} & \textbf{Value} \\
\midrule
Optimizer & AdamW \\
Learning rate & $1 \times 10^{-4}$ \\
Weight decay & 0.05 \\
Batch size & 64 \\
Epochs & 100 \\
LR scheduler & Linear decay (Trainer default) \\
Evaluation frequency & Every 500 steps \\
Precision & FP16 (CUDA) / FP32 (MPS) \\
Checkpoint selection & Best by validation accuracy \\
\bottomrule
\end{tabular}
\end{table}

A checkpoint (a full snapshot of the model weights, optimizer state, and training progress) is saved every 500 training steps, and the model is also evaluated on the held-out test set at each of these intervals. To limit disk usage, only the 3 most recent checkpoints are kept on disk at any time; when a fourth checkpoint is written the oldest is deleted. For example, after step 2000 the checkpoints for steps 1000, 1500, and 2000 would be on disk, while the checkpoint for step 500 would have been removed. At the end of training, the HuggingFace Trainer automatically loads the checkpoint with the highest top-1 validation accuracy (tracked via \texttt{eval\_accuracy} at each 500-step evaluation) and uses it as the final model for evaluation and release.

\subsection{Infrastructure}

Training is conducted using PyTorch with HuggingFace Transformers and the PEFT library for LoRA integration on Vast.ai cloud GPU instances equipped with NVIDIA RTX 3090 GPUs (24\,GB VRAM). Mixed-precision training (FP16) is enabled to reduce memory usage and training time. With the LoRA $r{=}8$ configuration and a batch size of 64, each epoch takes approximately 20 minutes; total training time for 100 epochs is approximately 33 hours per configuration.

\section{Results}

\subsection{Classification Accuracy}

The best-performing configuration (LoRA $r{=}8$) achieves \textbf{99.0\% top-1 accuracy} on the held-out test set across 394 font classes (Table~\ref{tab:severity}). LoRA $r{=}8$ achieved a peak validation accuracy of 99.0\% at epoch 99 after full 100-epoch training, marginally outperforming $r{=}16$ which peaked at 98.9\%. At a fixed 50-epoch budget (Table~\ref{tab:baselines}), the two configurations perform identically at 98.7\%. Wilson score 95\% confidence intervals (Table~\ref{tab:baselines}) confirm that the differences between LoRA ranks are not statistically significant, suggesting rank has minimal impact beyond $r{=}4$ for this task. Given the fine-grained nature of the task---many classes differ only in weight (e.g., Roboto-300 vs.\ Roboto-400) or subtle stylistic features---this represents strong performance.

\subsection{Analysis}

\begin{figure}[t]
    \centering
    \includegraphics[width=\textwidth]{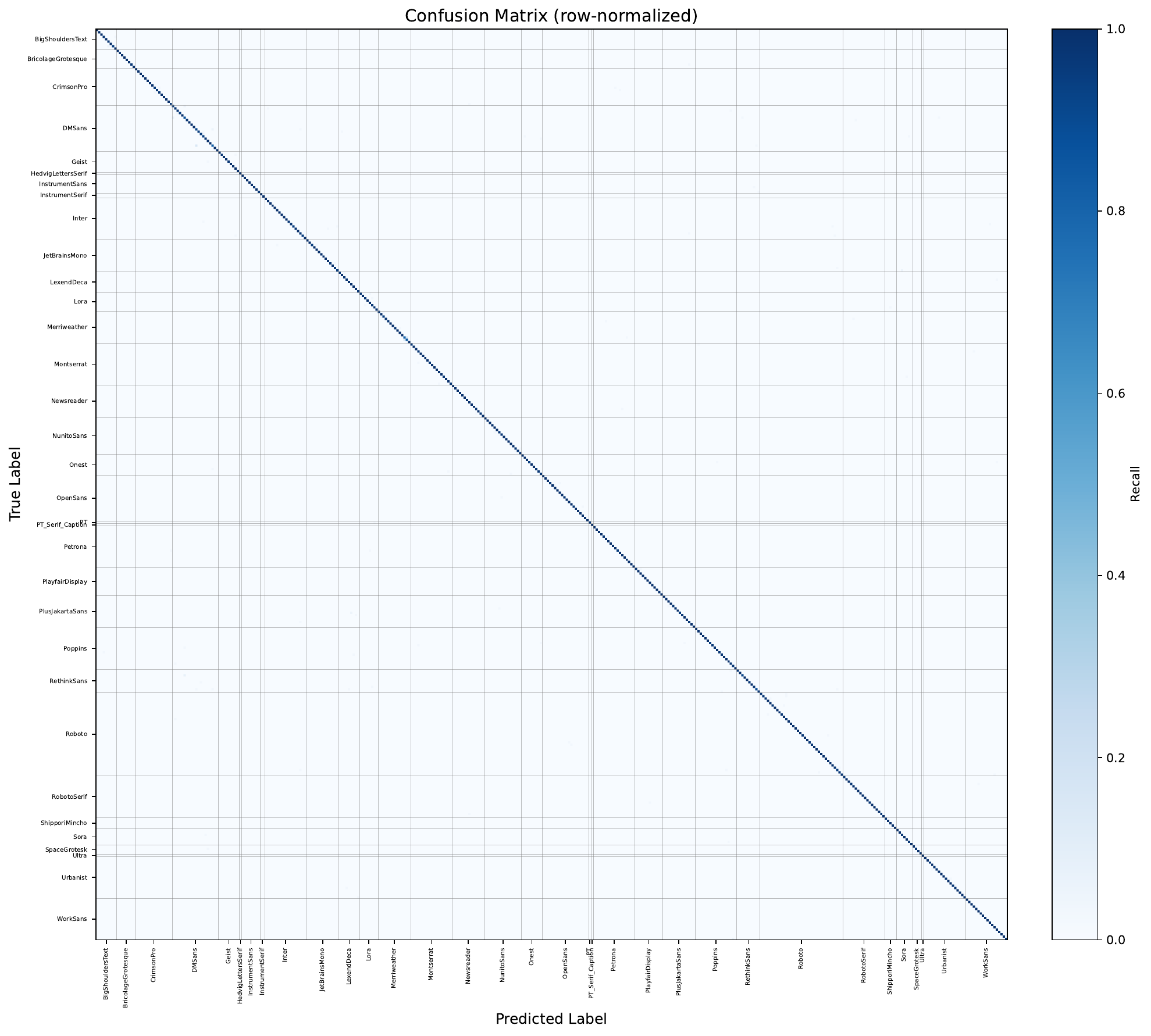}
    \caption{Row-normalized confusion matrix across all font classes, grouped by font family. The strong diagonal indicates high per-class accuracy, with remaining off-diagonal mass primarily within family blocks (weight variant confusion).}
    \label{fig:confusion_matrix}
\end{figure}

We compute a full confusion matrix across all classes to analyze per-class performance and systematic misclassification patterns. Figure~\ref{fig:confusion_matrix} shows the row-normalized confusion matrix with font families grouped together. Key observations include:

\begin{itemize}[leftmargin=*]
    \item \textbf{Weight variant confusion.} The most frequent individual error pairs remain within-family weight confusions (e.g., weight 400 vs.\ 500), though at 99\% accuracy the error distribution also includes cross-family pairs with similar visual characteristics. This is expected, as the visual difference between adjacent weights can be as small as a few pixels of stroke width at the $224 \times 224$ resolution.

    \item \textbf{Cross-family accuracy.} Classification between distinct font families (e.g., a serif vs.\ sans-serif) is near-perfect, indicating the model captures high-level typographic structure effectively.

    \item \textbf{Robustness to content.} Performance is consistent across sentence, number, currency, and percentage test images, suggesting the model learns font-intrinsic features rather than content-specific patterns.
\end{itemize}

\begin{figure}[t]
    \centering
    \includegraphics[width=0.85\textwidth]{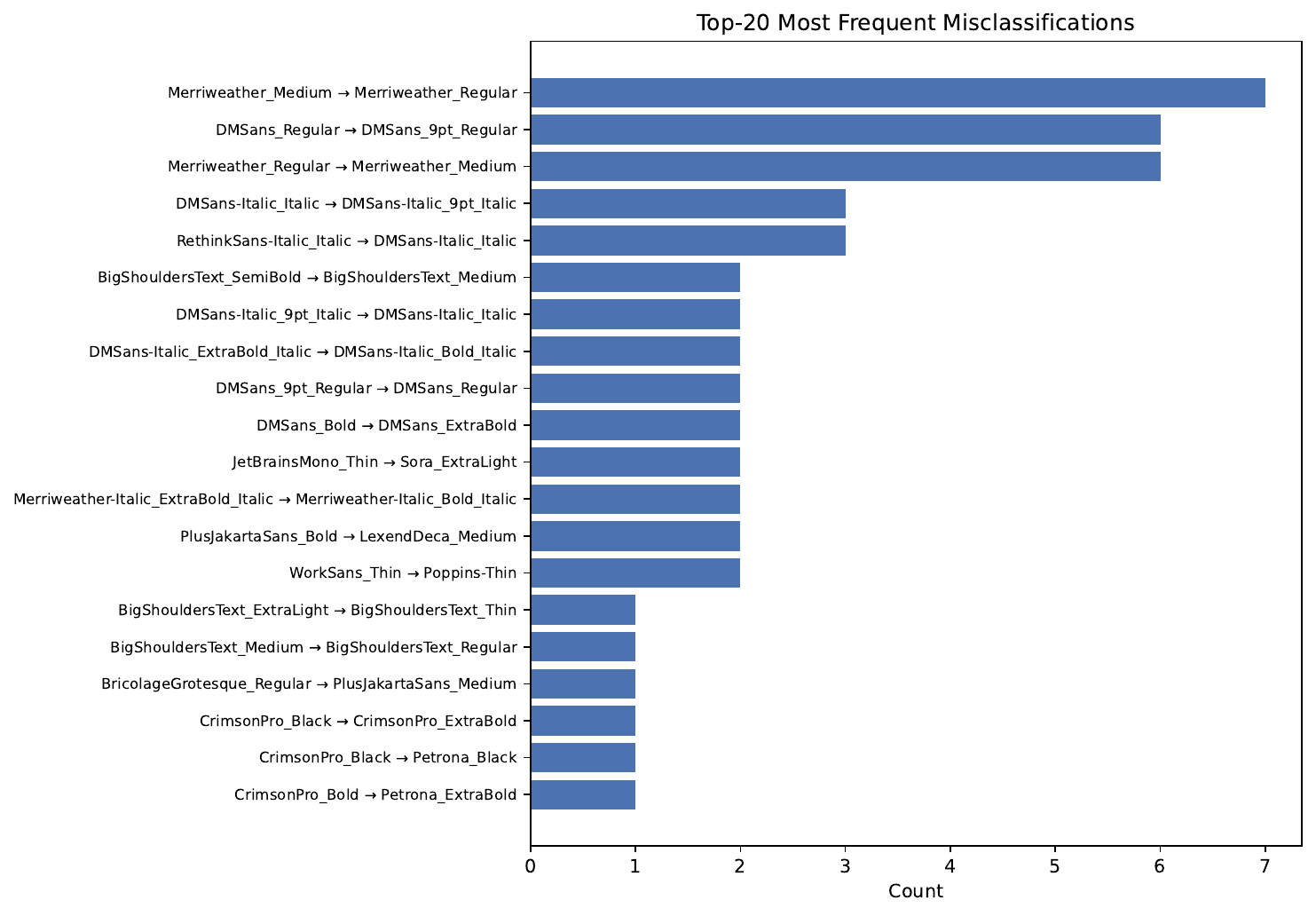}
    \caption{Top-20 most frequent misclassification pairs. At 99\% top-1 accuracy, the remaining misclassifications are diverse---including both within-family weight confusions and cross-family pairs with similar visual characteristics (e.g., thin sans-serif variants).}
    \label{fig:top_confused_pairs}
\end{figure}

\begin{figure}[t]
    \centering
    \includegraphics[width=\textwidth]{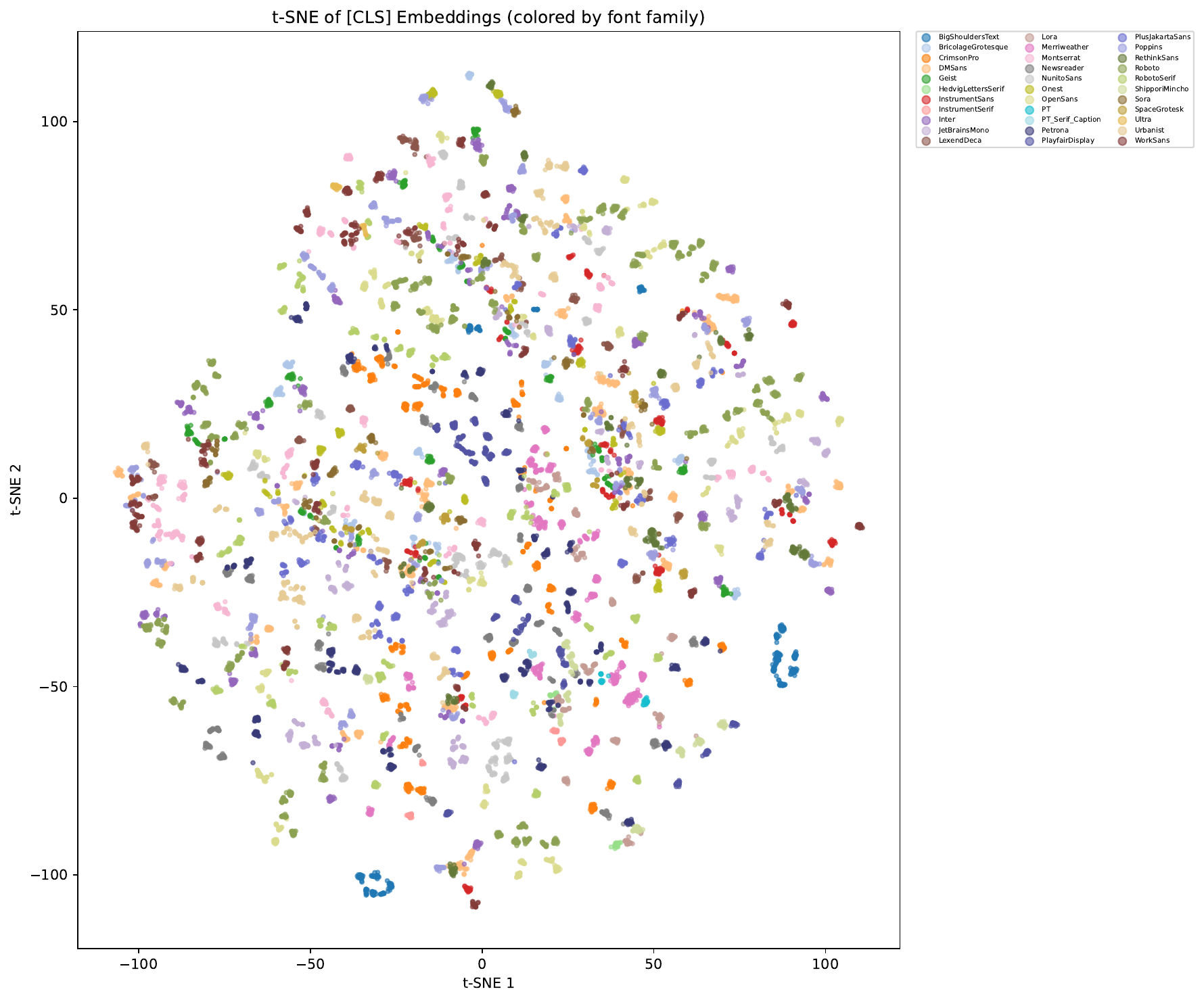}
    \caption{t-SNE visualization of \texttt{[CLS]} token embeddings from the final hidden layer, colored by font family. Points belonging to the same family cluster tightly together, indicating that the model learns a representation space that groups typographically related variants. Distinct font categories (serif, sans-serif, monospace) occupy well-separated regions.}
    \label{fig:tsne_embeddings}
\end{figure}

\begin{figure}[t]
    \centering
    \includegraphics[width=0.85\textwidth]{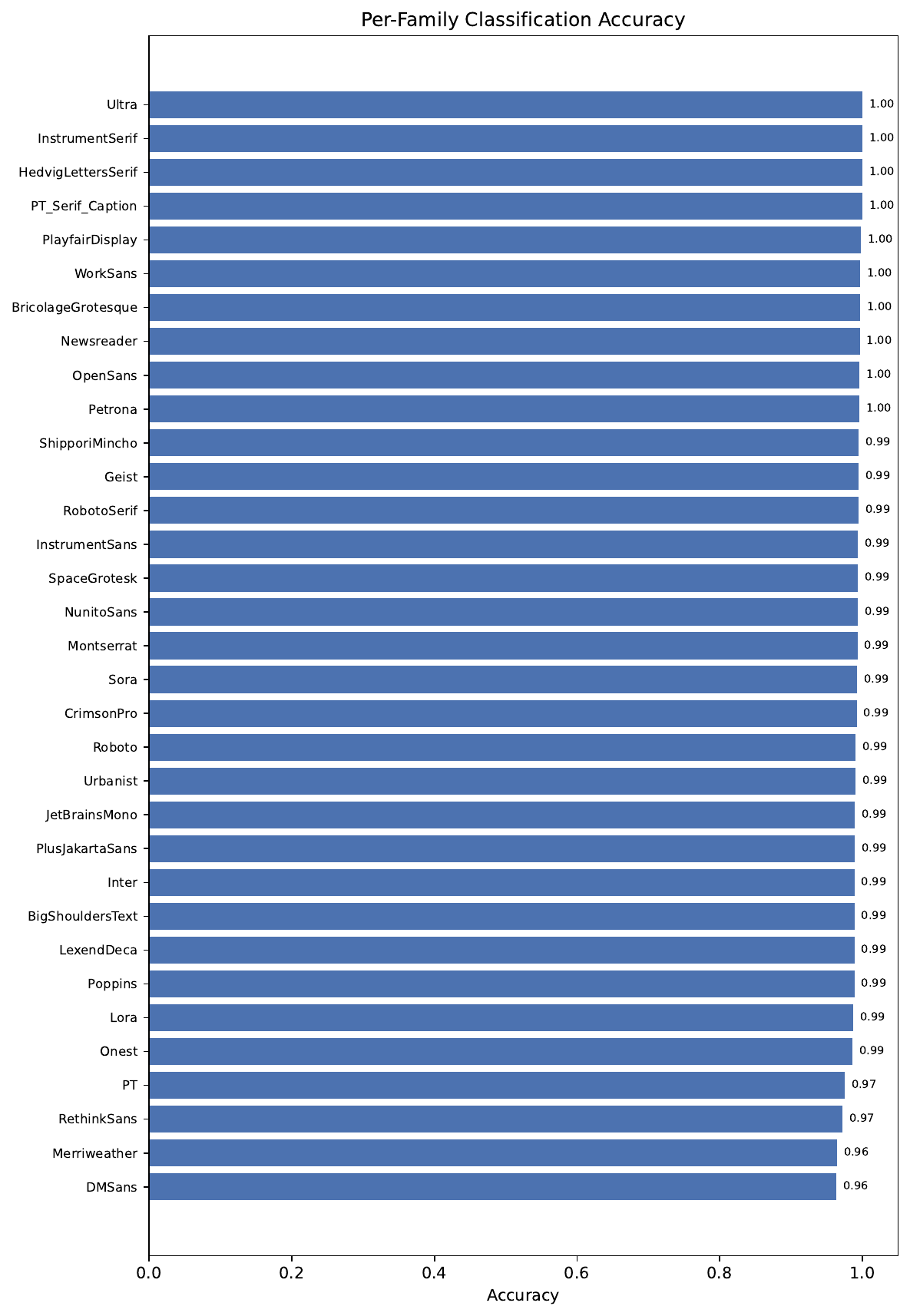}
    \caption{Top-1 variant-level accuracy broken down by font family, sorted from lowest to highest. Most families achieve near-perfect accuracy; lower-performing families tend to have many visually similar weight variants.}
    \label{fig:per_family_accuracy}
\end{figure}

\begin{figure}[t]
    \centering
    \includegraphics[width=0.75\textwidth]{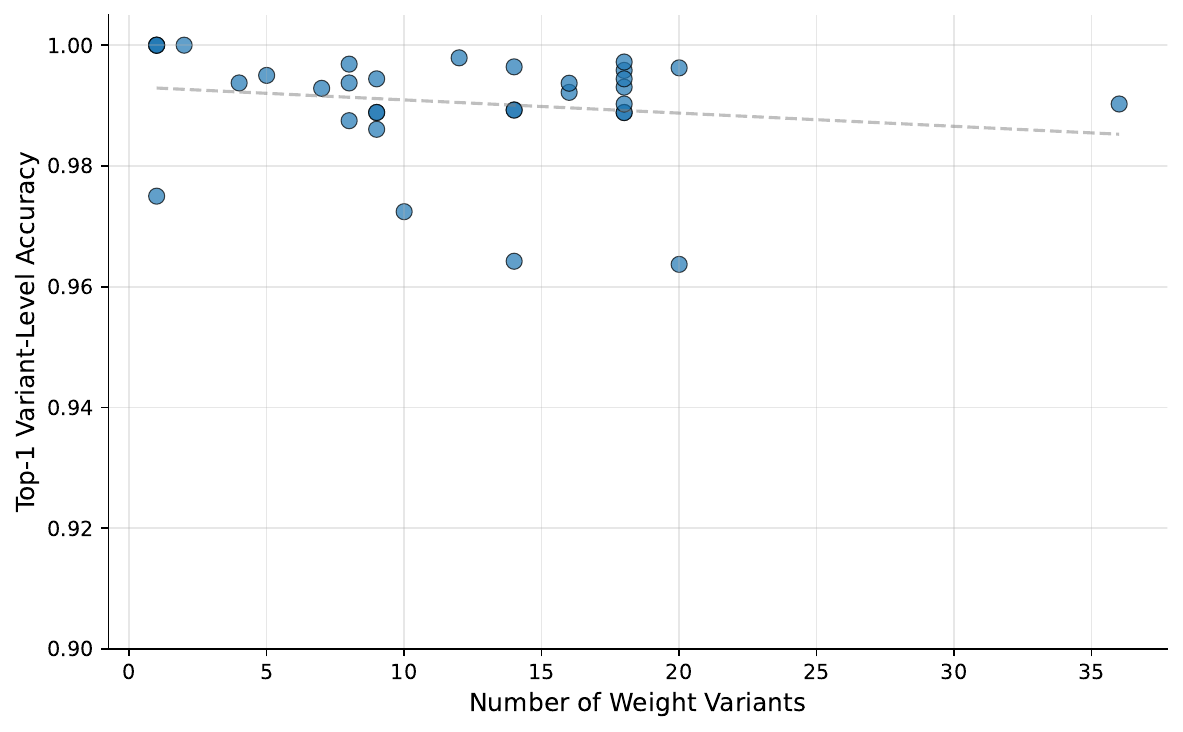}
    \caption{Top-1 variant-level accuracy plotted against the number of weight variants in each family. At 99\% overall accuracy, per-family accuracy is uniformly high (0.96--1.00), though a slight downward trend remains for families with many visually similar weight variants. Dashed line shows an OLS linear fit.}
    \label{fig:accuracy_vs_variants}
\end{figure}

\begin{figure}[t]
    \centering
    \includegraphics[width=0.85\textwidth]{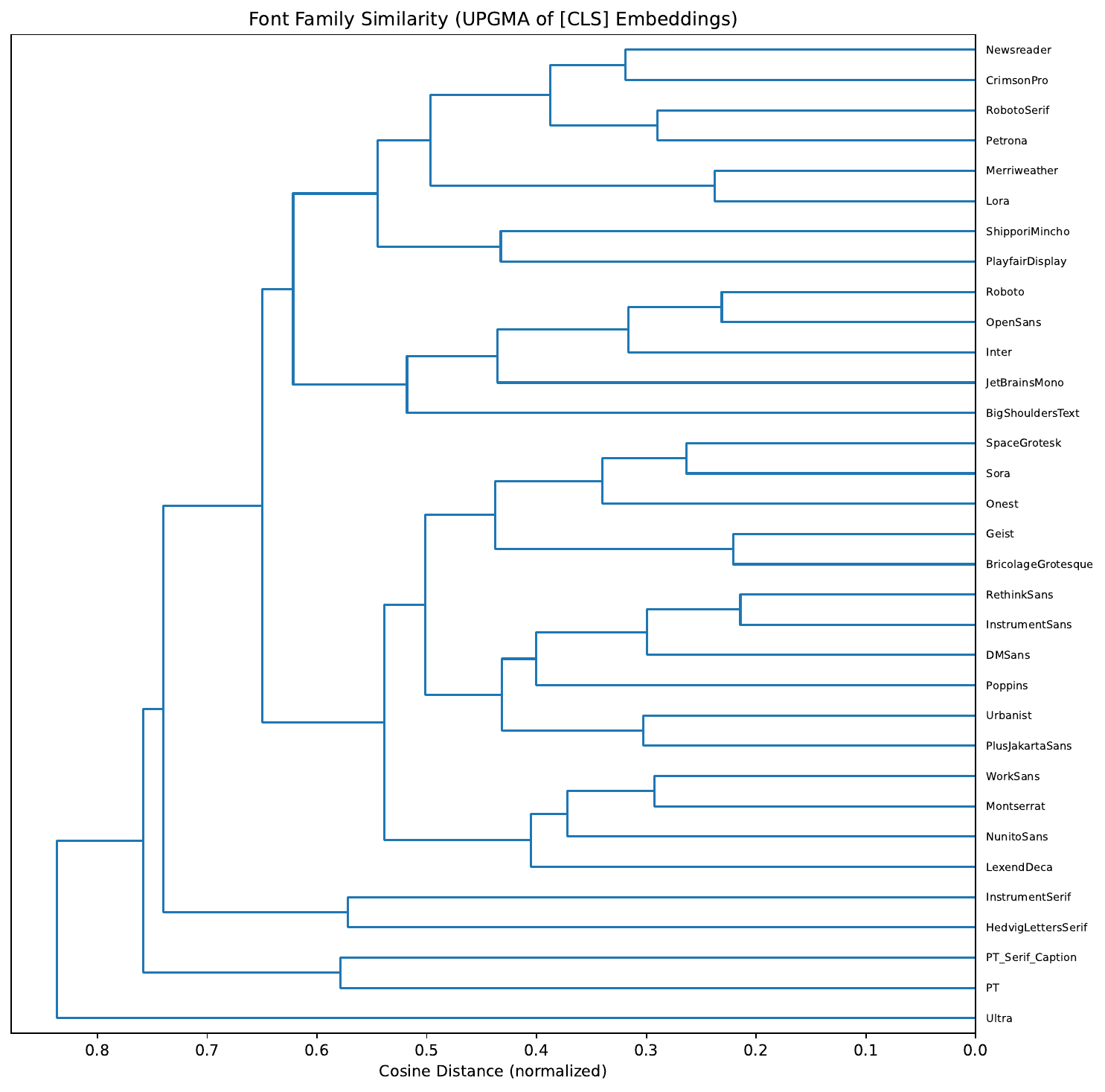}
    \caption{Hierarchical clustering (UPGMA) of 32 font families by cosine distance
    between mean \texttt{[CLS]} embeddings. The model's learned representations recover
    typographic structure: serif, sans-serif, monospace, and display families
    occupy distinct branches.}
    \label{fig:dendrogram}
\end{figure}

\paragraph{Severity-weighted error analysis.}
Top-1 accuracy treats all misclassifications equally, but confusing two adjacent weights of the same family (e.g., Roboto-400 vs.\ Roboto-500) is far less severe than confusing a serif with a sans-serif. We introduce two complementary metrics that capture this distinction.

\subparagraph{Family-level accuracy.}
The simplest correction: collapse all weight variants into their parent family and score a prediction as correct if the predicted and true labels share the same family, regardless of weight. The gap between variant-level and family-level accuracy directly quantifies how much of the raw error rate is due to within-family weight confusion versus genuine cross-family misidentification.

\subparagraph{Relative severity.}
For a continuous measure, we define a \emph{severity-weighted error rate} (SWER) using a typographic metadata distance that is independent of the model being evaluated. The pairwise distance $d(i, j)$ between font classes $i$ and $j$ is defined by a tier structure reflecting typographic similarity:
\begin{itemize}[leftmargin=*,nosep]
    \item $d = 0$: same variant (correct prediction).
    \item $d = 0.2 + 0.2 \times \min\!\bigl(\lvert w_i - w_j \rvert / 800,\; 1\bigr)$: same family, different weight, where $w_i, w_j \in \{100, 200, \ldots, 900\}$ are CSS font-weight values and the denominator 800 is the maximum possible weight difference (e.g., 400 vs.\ 500 yields $d = 0.225$; 100 vs.\ 900 yields $d = 0.4$).
    \item $d = 0.7$: different family, same typographic category (e.g., both sans-serif).
    \item $d = 1.0$: different typographic category (e.g., serif vs.\ sans-serif).
\end{itemize}
This distance function is symmetric and assigns zero distance only to identical variants and is fully determined by font metadata---it does not depend on the model's learned representations, avoiding the circularity that would arise from using the model's own embeddings to grade its errors.

For each prediction, the severity is $s(y, \hat{y}) = 0$ if correct and $d(y, \hat{y})$ otherwise. The SWER is the mean severity across all $N$ test predictions:
\begin{equation}
    \text{SWER} = \frac{1}{N} \sum_{n=1}^{N} s(y_n, \hat{y}_n)
\end{equation}
The absolute value of SWER depends on an arbitrary choice of distance normalization and is therefore not directly interpretable. However, the \emph{relative severity}---the ratio of the model's SWER to the expected SWER under uniform random predictions---is a dimensionless, scale-invariant quantity that cancels the normalization constant:
\begin{equation}
    \Pi = \frac{\text{SWER}}{\text{SWER}_{\text{random}}}, \qquad
    \text{SWER}_{\text{random}} = \frac{1}{K^2} \sum_{i,j} d(i,j)
\end{equation}
This ratio is invariant to any linear rescaling of the distance matrix: if all distances are multiplied by a constant $\alpha$, both numerator and denominator scale by $\alpha$ and the ratio is unchanged. A value of $\Pi = 0$ indicates a perfect classifier; $\Pi = 1$ indicates errors as severe as random guessing.

Table~\ref{tab:severity} reports both metrics. Family-level accuracy confirms that most errors are within-family weight confusions. The relative severity $\Pi$ provides a continuous, scale-invariant confirmation: the model's errors carry only a fraction of the cost expected from random predictions.

\begin{table}[h]
\centering
\caption{Error severity analysis. Family-level accuracy collapses weight variants; relative severity $\Pi$ is a scale-invariant ratio of the model's typographic-distance-weighted error to the random baseline.}
\label{tab:severity}
\begin{tabular}{lr}
\toprule
\textbf{Metric} & \textbf{Value} \\
\midrule
Top-1 accuracy (variant-level) & \rawAccuracy{} \\
\textbf{Family-level accuracy} & \textbf{\familyAccuracy{}} \\
\midrule
SWER (model) & \swerValue{} \\
SWER (random baseline) & \swerRandom{} \\
\textbf{Relative severity} $\Pi$ & \textbf{\swerRelative{}} \\
\bottomrule
\end{tabular}
\end{table}

Figure~\ref{fig:dendrogram} visualizes the font family similarity hierarchy recovered by UPGMA clustering on cosine distances between mean \texttt{[CLS]} embeddings. Note that this dendrogram uses model-derived distances (to show what the model has learned), whereas the SWER metric above uses the metadata-based typographic distance (to provide a model-independent evaluation). Serif, sans-serif, monospace, and display families occupy distinct branches in both, confirming that the model's learned representations align with typographic structure.

\subsection{Baseline Comparisons}

To contextualize the efficiency--accuracy trade-off of our LoRA approach, we compare against standard fine-tuning strategies applied to the same DINOv2-Base backbone and dataset. Table~\ref{tab:baselines} reports the results.

\begin{table}[h]
\centering
\caption{Comparison of fine-tuning strategies on DINOv2-Base with 394 font classes. All methods are evaluated at 50 training epochs using the same training data, hyperparameters (where applicable), and evaluation protocol.}
\label{tab:baselines}
\begin{tabular}{lrrrr}
\toprule
\textbf{Method} & \textbf{Trainable Params} & \textbf{Top-1 Acc} & \textbf{95\% CI} & \textbf{$\Pi$} \\
\midrule
ResNet-50 (CNN)      & 25.6M & 98.8\% & $\pm$0.17\% & 0.005 \\
\midrule
\multicolumn{5}{l}{\emph{DINOv2-Base backbone:}} \\
Linear Probe         & 606K  & 36.6\% & $\pm$0.75\% & 0.479 \\
LoRA ($r{=}4$)       & 753K  & 97.9\% & $\pm$0.21\% & 0.013 \\
LoRA ($r{=}8$)       & 900K  & 98.7\% & $\pm$0.16\% & 0.007 \\
LoRA ($r{=}16$)      & 1.2M  & 98.7\% & $\pm$0.16\% & 0.007 \\
Full Fine-Tuning     & 87.2M & 93.4\% & $\pm$0.31\% & 0.028 \\
\bottomrule
\end{tabular}
\end{table}

\begin{figure}[t]
    \centering
    \includegraphics[width=\textwidth]{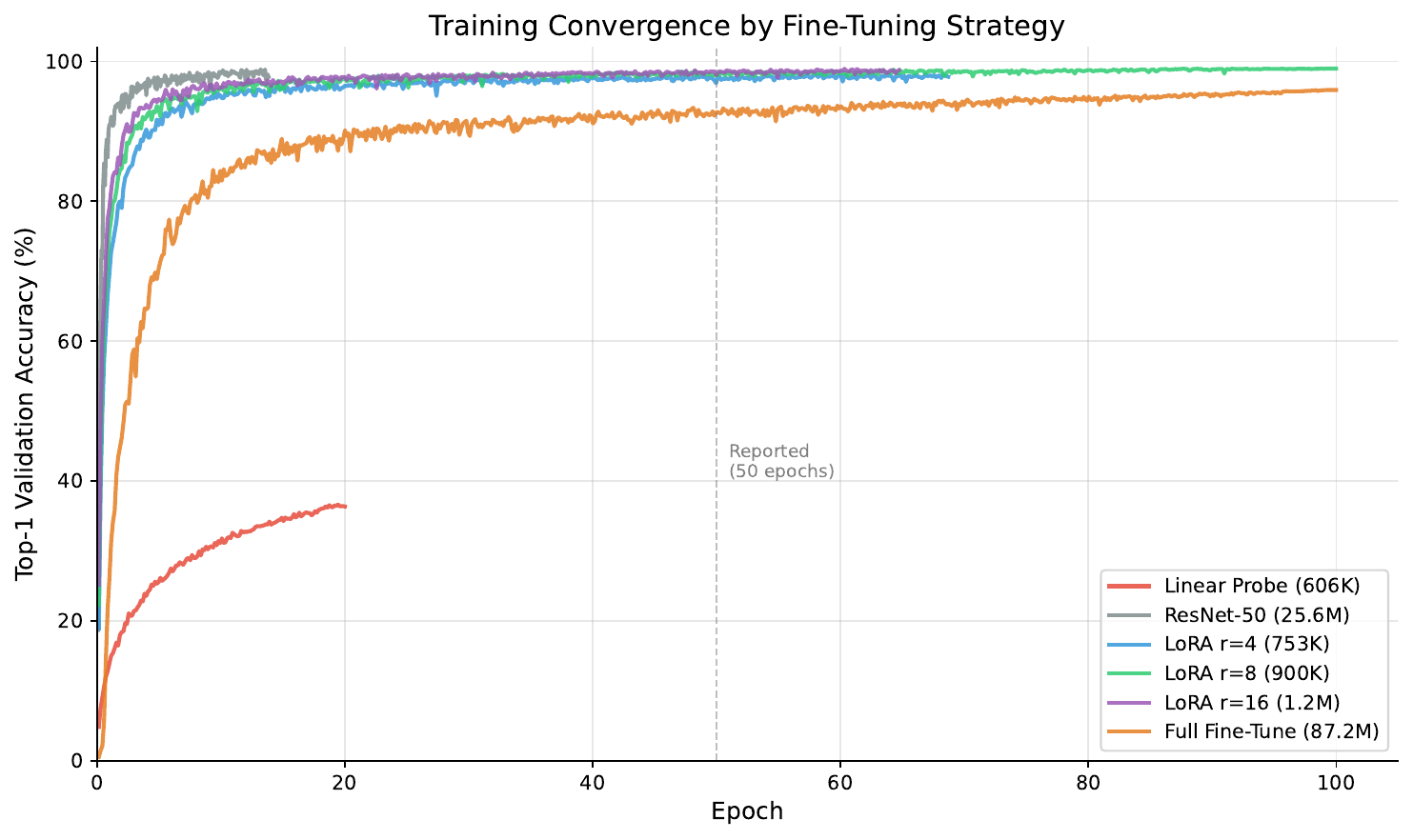}
    \caption{Validation accuracy over training epochs for each fine-tuning strategy. LoRA variants converge to $>$98\% within 10 epochs, while full fine-tuning requires $\sim$80 epochs to reach 96\%. The dashed line marks the 50-epoch evaluation point used in Table~\ref{tab:baselines}.}
    \label{fig:convergence}
\end{figure}

ResNet-50 achieves 98.8\% accuracy, matching the LoRA configurations despite having 25.6M trainable parameters. This is not surprising: the ResNet-50 baseline is fine-tuned from ImageNet-pretrained weights, so it also benefits from transfer learning rather than training from scratch. LoRA's advantage over ResNet-50 is therefore not raw accuracy but efficiency---LoRA trains 28$\times$ fewer parameters, produces a ${\sim}$3\,MB adapter file versus a ${\sim}$100\,MB full model, and converges faster (LoRA reaches 95\% accuracy by epoch 5, whereas ResNet-50 requires approximately 10 epochs; see Figure~\ref{fig:convergence}). These differences matter for deployment size, iteration speed, and the ability to maintain a single frozen backbone across multiple fine-tuned tasks.

\begin{figure}[t]
    \centering
    \includegraphics[width=\textwidth]{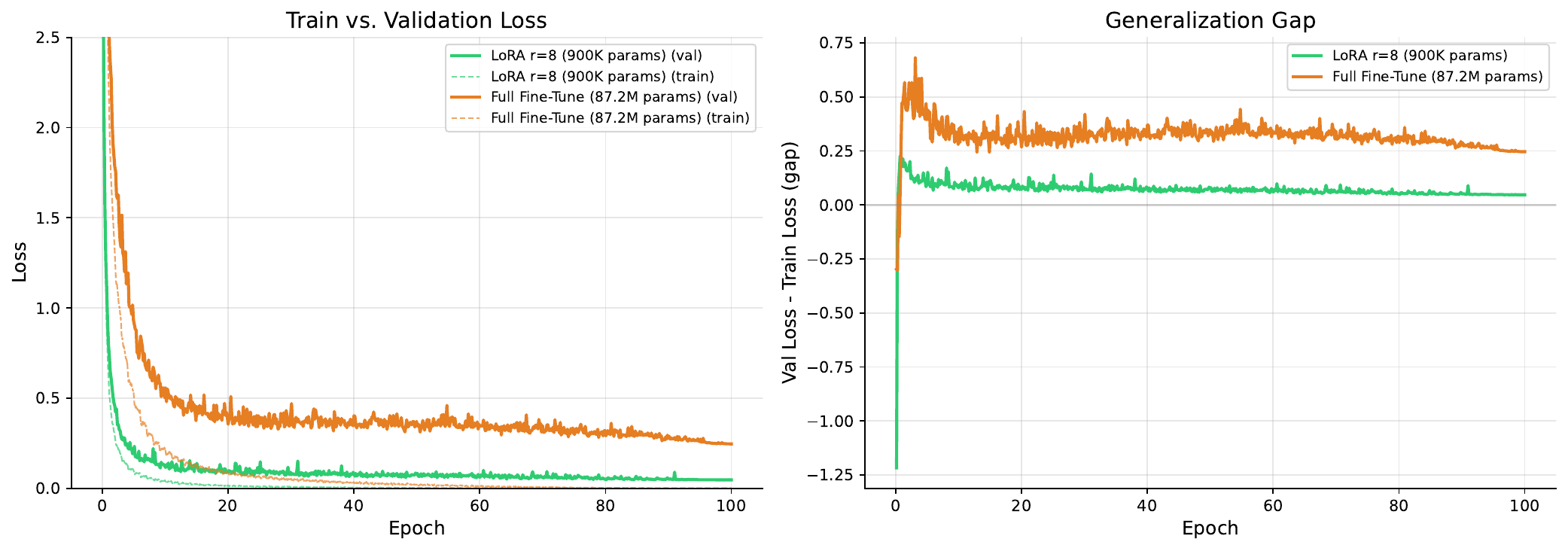}
    \caption{Left: training and validation loss for LoRA $r{=}8$ and full fine-tuning. Right: generalization gap (validation loss minus smoothed training loss). Full fine-tuning exhibits a widening gap indicative of overfitting, while LoRA maintains a narrow gap throughout training. The briefly negative gap early in LoRA training is expected: dropout is active during training but disabled during evaluation, so validation loss can temporarily fall below training loss.}
    \label{fig:overfitting}
\end{figure}

We attribute full fine-tuning's underperformance (93.4\% at 50 epochs, 95.9\% converged) to overfitting: with all 87.2M parameters trainable, the model has far more capacity than needed for ${\sim}$575 training images per class, leading to degradation of the pretrained features that LoRA preserves by construction. Figure~\ref{fig:overfitting} provides direct evidence: the generalization gap (validation loss minus training loss) widens substantially for full fine-tuning over the course of training, while LoRA $r{=}8$ maintains a narrow gap throughout.

Figure~\ref{fig:convergence} shows the validation accuracy over training for each method. The LoRA configurations converge rapidly, reaching 95\% accuracy within the first 5 epochs and plateauing near 99\% by epoch 30. Full fine-tuning converges more slowly, requiring approximately 80 epochs to reach its final accuracy of 95.9\%. The linear probe converges quickly but to a much lower ceiling, confirming that the frozen DINOv2 features alone are insufficient for fine-grained font discrimination. These convergence profiles demonstrate that LoRA's parameter efficiency translates directly to faster training convergence, not just fewer trainable parameters.

The SWER column in Table~\ref{tab:baselines} reveals a qualitative dimension beyond accuracy. The linear probe's $\Pi = 0.479$ indicates that its errors are nearly half as severe as random guessing---frozen DINOv2 features fail not just frequently but categorically, confusing fonts across typographic categories (e.g., serif with sans-serif). In contrast, LoRA's errors ($\Pi \approx 0.007$) are overwhelmingly within-family weight confusions, which are typographically benign. Full fine-tuning's errors ($\Pi = 0.028$) are 4$\times$ more severe than LoRA's, consistent with the overfitting hypothesis: degraded features produce less typographically coherent predictions.

\subsection{Generalization to Real-World Images}

Our model is trained and evaluated entirely on synthetically rendered text. The most established real-world font recognition benchmark, AdobeVFR \citep{wang2015deepfont}, contains 4,384 expert-labeled images spanning 617 font classes---however, these are drawn almost exclusively from commercial typefaces (Adobe Type Library, Linotype, Monotype), with zero overlap with our 32 Google Fonts families. Direct comparison is therefore not possible without retraining on a shared label space.

This highlights a gap in the font recognition literature: no public real-world benchmark exists for the open-source fonts that dominate modern web and mobile design. We are assembling a manually curated evaluation set of web typography screenshots for our target fonts, leveraging the fact that Google Fonts usage metadata identifies which websites use each family. Results on this real-world benchmark will be reported in a future revision. We note that our model is deployed in a production design tool and performs well on web screenshots in practice, though we refrain from reporting anecdotal production metrics as a substitute for controlled evaluation.

\section{Deployment}

The trained model is released on HuggingFace Hub as \texttt{dchen0/font\_classifier\_v4}.\footnote{\url{https://huggingface.co/dchen0/font_classifier_v4}} Before upload, the LoRA adapters are merged into the base weights (see Section~\ref{sec:lora}), so the published checkpoint requires no additional libraries beyond HuggingFace Transformers. A custom \texttt{handler.py} is included for HuggingFace Inference Endpoints, accepting raw image bytes or base64-encoded strings and returning top-5 predictions with softmax confidence scores.

\section{Discussion}

\paragraph{Strengths.} The combination of DINOv2's pre-trained representations and LoRA's parameter efficiency makes this approach practical for deployment. The synthetic data pipeline is fully reproducible and can be extended to new fonts without manual labeling. Built-in preprocessing eliminates a common source of train--serve skew.

\paragraph{Limitations.} The model is trained exclusively on synthetically rendered text and may not generalize perfectly to photographs of printed text, handwritten annotations overlaid on typed text, or heavily stylized graphics. The current system classifies images at the font level but does not localize text regions---it assumes the input is a cropped text region. Performance degrades on font weight variants that are visually near-identical at low resolution.

\paragraph{Future work.} Natural extensions include: (1) training on a broader set of fonts beyond the current 32 families, (2) incorporating real-world training images via domain adaptation, (3) jointly predicting font family and attributes (weight, width, style) in a multi-task framework, and (4) integrating text detection to enable end-to-end font recognition from uncropped images.

\section{Conclusion}

We have introduced GoogleFontsBench, a public benchmark for open-source font classification that fills a gap in existing evaluation resources dominated by commercial typefaces. On this benchmark, LoRA-adapted DINOv2 achieves 99.0\% top-1 accuracy while training only 1\% of parameters, with typographically mild errors ($\Pi = 0.007$). The benchmark, baseline results from six methods, evaluation protocol, and full training pipeline are publicly available to support further research in font recognition and fine-grained visual classification.\footnote{Source code: \url{https://github.com/Create-Inc/font-model}}

\section*{Acknowledgments}

We thank the Google Fonts team for maintaining the open-source font repository used to generate our training data.

\bibliographystyle{plainnat}

\end{document}